\documentclass[letterpaper]{article} 
\usepackage{aaai24}  
\usepackage{times}  
\usepackage{helvet}  
\usepackage{courier}  
\usepackage[hyphens]{url}  
\usepackage{graphicx} 
\usepackage{natbib}  
\usepackage{caption} 
\usepackage{algorithm}
\usepackage{algorithmic}
\usepackage{subcaption}

\usepackage{newfloat}
\usepackage{listings}
\usepackage{mhchem} 
\usepackage{amssymb}
\usepackage{dsfont}

\DeclareCaptionStyle{ruled}{labelfont=normalfont,labelsep=colon,strut=off} 
\floatstyle{ruled}
\newfloat{listing}{tb}{lst}{}
\floatname{listing}{Listing}
\title{Atom-Motif Contrastive Transformer for Molecular Property Prediction}
\author{
    Wentao Yu\textsuperscript{\rm 1},
    Shuo Chen\textsuperscript{\rm 2},
    Chen Gong\textsuperscript{\rm 1},
    Gang Niu\textsuperscript{\rm 2},
    Masashi Sugiyama\textsuperscript{\rm 2, 3}
}
\affiliations{
    \textsuperscript{\rm 1}Nanjing University of Science and Technology\\
    \textsuperscript{\rm 2}RIKEN\\
    \textsuperscript{\rm 3}The University of Tokyo\\

    \{wentao.yu, chen.gong\}@njust.edu.cn; shuo.chen.ya@riken.jp; gang.niu.ml@gmail.com; sugi@k.u-tokyo.ac.jp
}

\usepackage{bibentry}

\begin{document}

\maketitle

\begin{abstract}
Recently, Graph Transformer (GT) models have been widely used in the task of Molecular Property Prediction (MPP) due to their high reliability in characterizing the latent relationship among graph nodes (\textit{i.e.}, the atoms in a molecule). However, most existing GT-based methods usually explore the basic interactions between pairwise atoms, and thus they fail to consider the important interactions among critical motifs (\textit{e.g.}, functional groups consisted of several atoms) of molecules. As motifs in a molecule are significant patterns that are of great importance for determining molecular properties (\textit{e.g.}, toxicity and solubility), overlooking motif interactions inevitably hinders the effectiveness of MPP. To address this issue, we propose a novel Atom-Motif Contrastive Transformer (AMCT), which not only explores the atom-level interactions but also considers the motif-level interactions. Since the representations of atoms and motifs for a given molecule are actually two different views of the same instance, they are naturally aligned to generate the self-supervisory signals for model training. Meanwhile, the same motif can exist in different molecules, and hence we also employ the contrastive loss to maximize the representation agreement of identical motifs across different molecules. Finally, in order to clearly identify the motifs that are critical in deciding the properties of each molecule, we further construct a property-aware attention mechanism into our learning framework. Our proposed AMCT is extensively evaluated on seven popular benchmark datasets, and both quantitative and qualitative results firmly demonstrate its effectiveness when compared with the state-of-the-art methods.
\end{abstract}

\section{Introduction}
\begin{figure}[t]
    \centering
    \includegraphics[width=1\columnwidth]{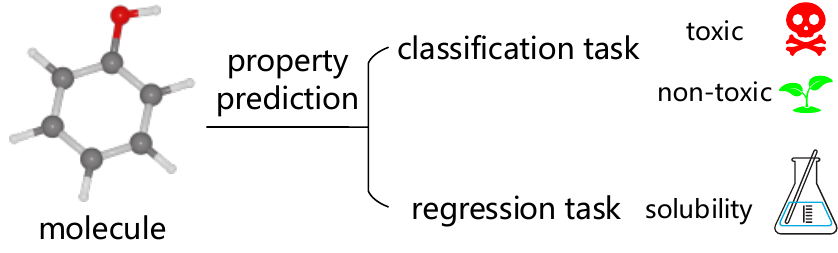}
    \caption{Molecular property prediction can be formulated as a classification task or a regression task, depending on the specific prediction objective.}
    \label{fig0}
\end{figure}

Molecular Property Prediction (MPP) is a fundamental and critical task in many areas of basic science, such as chemistry, biomedicine, and pharmacology~\cite{LI2022103373}. The primary objective of MPP is to accurately predict the potential properties (\textit{e.g.}, toxicity and solubility) of a given molecule, and this prediction task is particularly challenging due to the intricate nature of interactions within molecules. As shown in Fig.~\ref{fig0}, depending on the specific prediction objective, MPP can be formulated as a regular classification task, where the goal is to assign a molecule to its predefined classes (\textit{e.g.}, toxic and non-toxic), or it can also be viewed as a regression task, aiming to predict continuous values (\textit{e.g.}, solubility) associated with the molecular properties.

\begin{figure}[t]
    \centering
    \includegraphics[width=0.7\columnwidth]{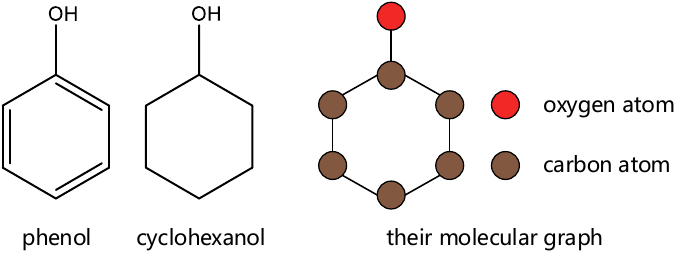}
    \caption{Molecular formulae of phenol and cyclohexanol, as well as their corresponding molecular graph. As the hydrogen atom is neglected in the MPP task, phenol and cyclohexanol have the same molecular graph.}
    \label{fig1}
\end{figure}

In recent years, Graph Transformer (GT) models have been widely used for the MPP task due to their high reliability in characterizing the latent relationship among atoms~\cite{NEURIPS2021_f1c15925, NEURIPS2022_5d4834a1, Wu_Fang_2023}. However, most existing GT-based methods usually explore the basic interactions between pairwise atoms, so they fail to consider the important interactions among critical motifs (\textit{e.g.}, functional groups consisted of several atoms) of molecules. Consequently, these methods may fail to recognize critical patterns hidden in motifs and thus they are limited in their abilities to represent molecules. This is attributed to the fact that motifs represent meaningful subgraph patterns that consistently emerge with notable frequencies~\cite{science2985594824} and play an even more fundamental role in determining molecular properties when compared with atoms~\cite{zhang2021motif, Wu_Fang_2023}. Furthermore, atom-level interactions are not always sufficient to predict molecular properties, and motif-level interactions are sometimes more important because they provide the detailed structural information of a molecule. For example, if we only consider atom-level interactions, phenol and cyclohexanol will have similar chemical properties, because they have the same molecular graph as shown in Fig.~\ref{fig1}. However, phenol is actually much more acidic than cyclohexanol. This is because the hydroxyl group in phenol interacts with the $\pi$ electron cloud on the benzene ring~\cite{ja2069592}, making it easier for phenol to release protons than cyclohexanol (we provide more details and present the motif interaction in phenol in the supplementary material).

In light of the above-mentioned considerations, we propose a novel Atom-Motif Contrastive Transformer (AMCT), which not only explores the atom-level (\textit{a.k.a.} low-level) interactions but also considers the motif-level (\textit{a.k.a.} high-level) interactions. To incorporate two levels of interactions and enhance the molecular representation capability, we propose to build the atom-motif contrastive learning (as shown in Fig.~\ref{fig2}), which is inspired by the popular work SimCLR~\cite{pmlrv119chen20j} in self-supervised learning. First, considering that the representations of atoms and motifs for a given molecule are actually two different views of the same instance, they are naturally aligned during the training process. As such, they can jointly provide the self-supervisory signals and thereby improve the reliability of the learned molecular representation. Second, it is revealed that identical motifs across different molecules usually have similar chemical properties~\cite{hawker2005convergence, smith2018infrared}. For example, carbon rings with \ce{NO2} or \ce{NH2} functional groups tend to be mutagenic~\cite{debnath1991structure, NEURIPS2020_e37b08dd}. It means that identical motifs should have consistent representations across all molecules. Therefore, we employ another contrastive loss to maximize the representation agreement of identical motifs across different molecules and thus obtaining distinguishable motif representations.

Moreover, in order to clearly identify the motifs that are critical in deciding the properties of each molecule, we further construct a property-aware attention mechanism (see the right panel of Fig.~\ref{fig2}) by using the cross-attention module. Specifically, the cross-attention module calculates the cross-attention weights between the molecular property embeddings and the motif representations. As a result, we can identify influential motifs based on the cross-attention weights. To the best of our knowledge, this is the first attempt to apply a property-aware attention mechanism in the GT model. Thanks to the effective atom-motif contrastive learning and the additional property-aware attention mechanism, the proposed AMCT can successfully extract informative features from molecules, therefore improving the performances in various MPP tasks. We compare our AMCT with eleven existing methods on seven popular benchmark datasets, and both quantitative and qualitative results clearly demonstrate the superiority of our proposed method. To sum up, the contributions of our work are as follows:
\begin{itemize}
    \item We propose a new AMCT to simultaneously explore the atom-level interactions and the motif-level interactions, so that we can successfully recognize critical patterns hidden in motifs and improve the reliability of MPP.
    \item We are the first to construct a property-aware attention mechanism in the GT model to identify the motifs that are critical in deciding the properties of each molecule.
    \item The experimental results on seven popular datasets firmly demonstrate the superiority of our proposed AMCT when compared with the state-of-the-art methods.
\end{itemize}

\section{Related Work}
This section reviews the typical works related to this paper, including MPP, GT model, and motif learning techniques.

\subsection{Molecular Property Prediction}
In the past decades, numerous methods have been proposed for the MPP task. Initially, these methods mainly depend on hand-crafted features such as molecular descriptors~\cite{todeschini2008handbook} and molecular fingerprints~\cite{rogers2010extended}. However, these approaches may suffer from limited scalability and flexibility~\cite{schwaller2021mapping, bbab152}. To address these challenges, deep learning-based approaches, such as GNNs and GT models, have been introduced to generate highly expressive molecular representations. For instance, Zhang~\textit{et al.}~\nocite{zhang2021motif}(2021b) pre-trained a GNN with a motif-based generation task. Although GNN-based approaches have shown promising results, they still have some limitations inherited from GNNs, such as over-smoothing~\cite{Li_Han_Wu_2018, Chen_Lin_Li_Li_Zhou_Sun_2020}, over-squashing~\cite{topping2022understanding}, and limited expressiveness~\cite{Morris_Ritzert_Fey_Hamilton_Lenssen_Rattan_Grohe_2019}. To deal with these deficiencies, more GT-based MPP methods have been proposed. For example, Ying~\textit{et al.}~\nocite{NEURIPS2021_f1c15925}(2021) introduced the structural information from graphs into the classical transformer~\cite{vaswani2017attention} and achieved promising results. Despite the noticeable achievements of GT-based MPP methods in recent years, most of them ignore the critical interactions among motifs. Therefore, in this paper, we explore the atom-level and motif-level interactions, simultaneously.

\subsection{Graph Transformer}
In contrast to the message-passing mechanism in a GNN, which primarily aggregates local neighborhood information, a GT model possesses the ability to capture interactions between each pair of nodes through the self-attention mechanism~\cite{pmlr-v162-chen22r}. As an early attempt to generalize the transformer to the graph case, Dwivedi and Bresson~\nocite{dwivedi2020generalization} (2021) employed Laplacian eigenvectors as positional encodings of nodes and computed the corresponding attention weights based on the neighborhood connectivity of each node. To enhance the representation ability of GT models in a self-supervised manner, Zhang~\textit{et al.}~\nocite{10004569}(2022) proposed to directly contrast two different graph views of the same instance, which shares the similar motivation with the typical works such as SimCLR~\cite{pmlrv119chen20j} and MoCo~\cite{He_2020_CVPR} in self-supervised learning. Due to the success of contrastive learning in visual tasks, lots of graph contrastive learning methods have been proposed such as GCA~\cite{1011453442381}. Contrastive learning has also been introduced into the MPP task. For example, Fang~\textit{et al.}~\nocite{Fang_Zhang_Y2022} (2022) contrasted molecular graphs generated by the knowledge-guided graph augmentations. To incorporate two levels of interactions, we explore the self-supervision by our proposed atom-motif contrast in this paper.


\subsection{Motif Learning}
Since motifs are significant subgraph patterns that appear consistently with remarkable frequencies~\cite{science2985594824}, they usually play a vital role in graph learning. Yu and Gao~(2022) constructed a heterogeneous motif graph containing both motif nodes and molecular nodes. They then used the message-passing mechanism to learn the heterogeneous motif graph. Similarly, Wu \textit{et al.}~(2023) proposed a heterogeneous molecular graph transformer called Molformer integrating both the atom-level and motif-level nodes. Nevertheless, Molformer highly depends on the exact 3D coordinates of each molecule, which is usually quite hard to acquire. Furthermore, feeding two types of nodes into the same self-attention module poses a practical challenge in distinguishing interactions between different node types. In comparison, we separate atom-level and motif-level interactions by using the atom encoder and motif encoder, respectively. Meanwhile, we build the self-supervision of atom-motif contrast, so that we can conduct the representation learning without using any 3D coordinate information. Consequently, our proposed method actually has remarkable distinctions and superiorities when compared with Molformer.

\section{Our Proposed Method}
In this section, we will discuss the main technical details of our proposed AMCT, including atom and motif encoding, property-aware decoding, and our proposed loss functions.
\subsection{Pipeline of Our Proposed Method}
\begin{figure*}[t]
    \centering
    \includegraphics[width=1.8\columnwidth]{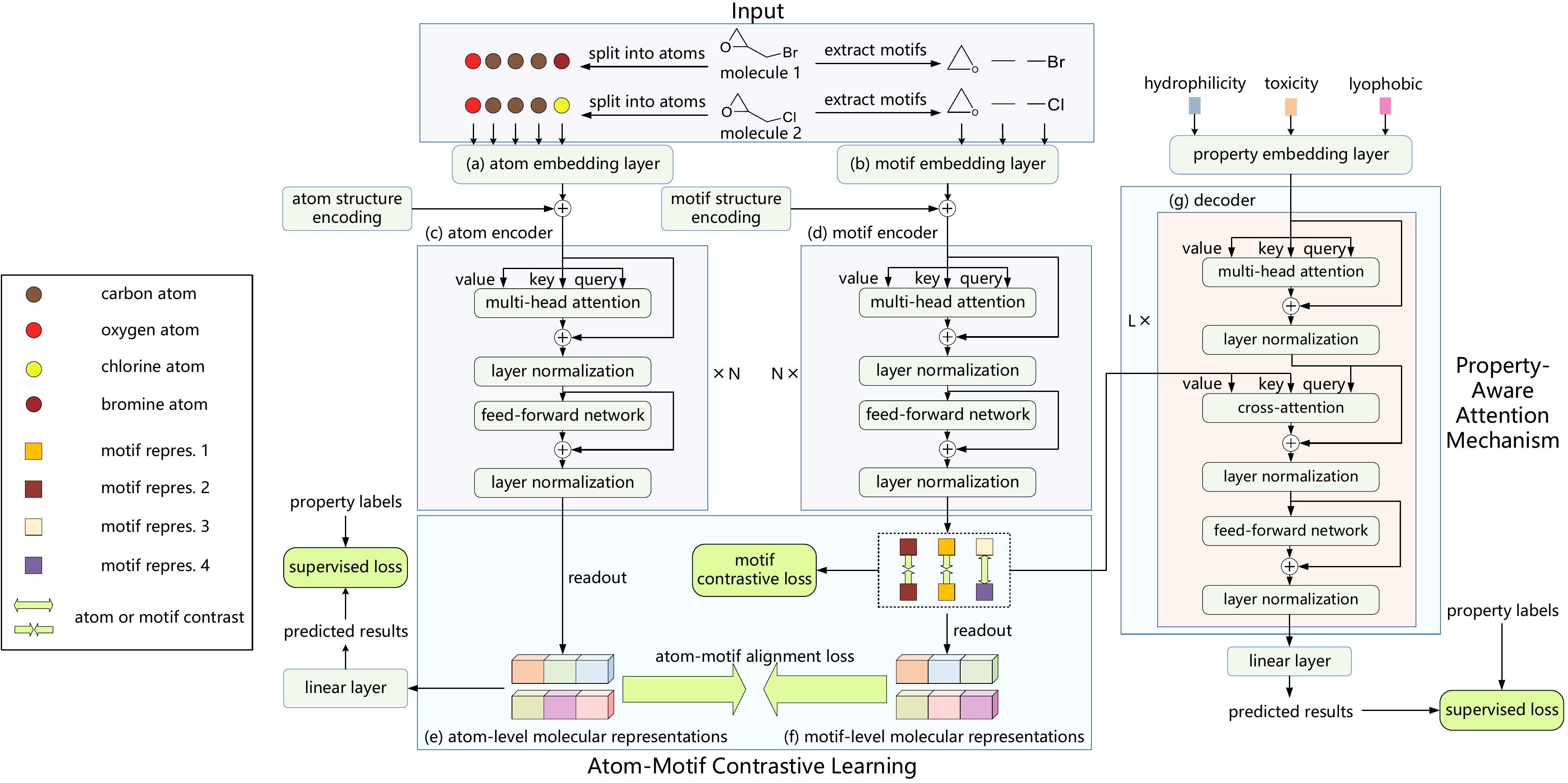}
    \caption{The framework of our proposed AMCT. First, two molecules are split into multiple atoms and segmented into multiple motifs, respectively. Second, the atom and motif embedding layers (\textit{i.e.}, (a) and (b)) generate atom and motif embeddings, respectively. Third, the atom and motif encoders (\textit{i.e.}, (c) and (d)) are used to obtain atom-level and motif-level molecular representations (\textit{i.e.}, (e) and (f)), respectively. Finally, the decoder (g) and a linear layer are invoked to obtain predicted results, where the atom-motif alignment loss, motif contrastive loss, and supervised losses are used for model training.}
    \label{fig2}
\end{figure*}

The framework of AMCT is shown in Fig.~\ref{fig2}. Given a batch of molecules, they are first split into a set of atoms and segmented into a set of motifs, respectively. Second, the atom embedding layer (Fig.~\ref{fig2}(a)) and the motif embedding layer (Fig.~\ref{fig2}(b)) generate atom embeddings and motif embeddings, respectively. Third, the atom encoder (Fig.~\ref{fig2}(c)) and the motif encoder (Fig.~\ref{fig2}(d)) are used to obtain atom-level and motif-level molecular representations (Fig.~\ref{fig2}(e) and Fig.~\ref{fig2}(f)), respectively. Finally, the decoder (Fig.~\ref{fig2}(g)) and a linear layer are invoked to obtain the predicted results, where the atom-motif alignment loss, motif contrastive loss, and supervised losses are utilized for model training.

\subsection{Atom Encoding}
In the process of atom encoding, we first obtain atom embeddings. Then, we use the atom encoder to extract atom-level interactions. Finally, atom-level molecular representations are obtained after the readout operation.

\subsubsection{Atom Embedding and Structure Encoding}
Here we adopt the atom embedding method provided by the Open Graph Benchmark (OGB)~\cite{hu2020open}, which is widely acknowledged in the field of MPP. After atom embedding, in a molecule with $n$ atoms, the $i$-th atom is represented by a feature vector $\mathbf{x}_i \in \mathbb{R}^{d}$, where $i=1,\ldots, n$ and $d$ is the dimensionality of $\mathbf{x}_i$. Meanwhile, we use the degree centrality to encode the structural information among atoms, namely the connective relationship of atoms in the molecule. Therefore, they can be represented as $\mathbf{b}_i \in \mathbb{R}^{d}$ for $i=1,\ldots, n$. Since the degree centrality is applied to each atom, we simply add it to the atom embedding as in Graphormer~\cite{NEURIPS2021_f1c15925}. Finally, given a molecule with $n$ atoms, the combination of atom embedding and structural information for the $i$-th atom is represented as $\mathbf{h}_i=\mathbf{x}_i+\mathbf{b}_i$ ($i=1,\ldots, n$), where $\mathbf{h}_i$ will be fed into the atom encoder.


\subsubsection{Atom Encoder and Readout}
The atom encoder (see Fig.~\ref{fig2}(c)) aims to capture the interactions among atoms. Given an input sequence of $\mathbf{h}_1, \ldots, \mathbf{h}_n$, we can represent them as a matrix $\mathbf{H} \in \mathbb{R}^{n \times d}$ with each row corresponding to an $\mathbf{h}_i$ ($i=1,\ldots, n$). The computation of Multi-Head Attention (MHA) in the atom encoder can be expressed as
\begin{equation}\label{eq1}
\left\{
    \begin{aligned}
    & \text{MultiHead}^\text{atom}(\mathbf{H})=\text{Con}(\text{head}_1^\text{atom}, \ldots, \text{head}_h^\text{atom})\mathbf{W}^{\mathbf{O}}, \\
    & \text{head}_i^\text{atom}=\text{Attention}(\mathbf{H}\mathbf{W}^{\mathbf{Q}}_i, \mathbf{H}\mathbf{W}^{\mathbf{K}}_i, \mathbf{H}\mathbf{W}^{\mathbf{V}}_i), \\
    & \text{Attention}(\mathbf{Q}, \mathbf{K}, \mathbf{V})=\text{softmax}\left(\mathbf{Q}\mathbf{K}^{\top }/\sqrt{d_k}\right)\mathbf{V}, \\
    \end{aligned}
\right.
\end{equation}
where $\mathbf{W}^{\mathbf{Q}}_i, \mathbf{W}^{\mathbf{K}}_i, \mathbf{W}^{\mathbf{V}}_i \in \mathbb{R}^{d \times d_k}$, and $\mathbf{W}^{\mathbf{O}} \in \mathbb{R}^{h d_k \times d}$ are projection matrices. The notations $\mathbf{Q}$, $\mathbf{K}$, and $\mathbf{V}$ represent the matrices of query, key, and value, respectively. Here $h$ is the number of attention heads, and $d_k=d/h$ denotes the number of columns of $\mathbf{W}^{\mathbf{Q}}_i$, $\mathbf{W}^{\mathbf{K}}_i$, and $\mathbf{W}^{\mathbf{V}}_i$. Moreover, `$\text{Con}$' denotes the concatenation operation performed along the column. By sequentially passing through $N$ layers, we acquire the atom representations. After that, we use a readout operation (namely, summation function) that aggregates the individual atom representations to generate the entire molecular graph representation (\textit{i.e.}, the atom-level molecular representation $\mathbf{h}^{\text{readout}} \in \mathbb{R}^{d}$).

\subsection{Motif Encoding}
Atom interactions successfully capture the low-level details, yet they ignore the high-level structural information among different molecules, so they are not sufficient to predict molecular properties in some cases (\textit{e.g.}, the example in Fig.~\ref{fig1}). Therefore, motif-level interactions are naturally introduced in our proposed AMCT. In the process of motif encoding, we first extract motifs and then obtain motif embeddings. After that, we use the motif encoder to extract motif-level interactions. Finally, motif-level molecular representations are obtained after the readout operation. Meanwhile, we use the atom-motif alignment loss and the motif contrastive loss to generate the self-supervisory signals.

\subsubsection{Motif Extraction}
We employ the tree decomposition of molecules algorithm~\cite{pmlrv80jin18a} to extract motifs. After motif extraction, a motif vocabulary is constructed by preprocessing all molecules in the dataset, which is presented in the supplementary material.


\subsubsection{Motif Embedding and Structure Encoding}
Here we use neural embedding vectors~\cite{pmlrv80jin18a} to represent each motif. This is analogous to using word embeddings to represent words in natural language processing tasks. After motif embedding, in a molecule with $m$ motifs, the $i$-th motif is represented by a feature vector $\mathbf{t}_i \in \mathbb{R}^{d}$, where $i=1,\ldots, m$. Meanwhile, we use the degree centrality to encode the structural information among motifs, namely the connective relationship of motifs in the molecule. Therefore, they can be represented as $\mathbf{e}_i \in \mathbb{R}^{d}$ for $i=1,\ldots, m$. Finally, given a molecule with $m$ motifs, the combination of motif embedding and structural information for the $i$-th motif is represented as $\mathbf{z}_i=\mathbf{t}_i+\mathbf{e}_i$ ($i=1,\ldots, m$), where $\mathbf{z}_i$ will be fed into the motif encoder.


\subsubsection{Motif Encoder and Readout}
The motif encoder (see Fig.~\ref{fig2}(d)) is designed to capture the interactions among motifs. Given an input sequence of $\mathbf{z}_1, \ldots, \mathbf{z}_m$, we can represent them as a matrix $\mathbf{Z} \in \mathbb{R}^{m \times d}$ with each row corresponding to a $\mathbf{z}_i$ ($i=1,\ldots, m$). By sequentially passing through $N$ layers, we acquire the motif representations denoted as $\mathbf{Z}^{\text{motif}} \in \mathbb{R}^{m \times d}$, where each row denotes the representation of each motif. After the summation function for the readout operation, we obtain the motif-level molecular representation $\mathbf{z}^{\text{readout}} \in \mathbb{R}^{d}$ from $\mathbf{Z}^{\text{motif}}$.

\subsubsection{Atom-Motif Alignment Loss}
To explore the additional new supervisory information provided by the motifs, we consider the similarity relationship between the atom-level and motif-level molecular representations. Since the representations of atoms and motifs for a given molecule are actually two different views of the same instance, they are naturally aligned to generate the self-supervisory signals for model training. Here, in a batch of $q$ molecules, the atom-motif alignment loss can be calculated as
\begin{equation}\label{eq7}
\mathcal{L}_\text{align}=\frac{1}{q}\sum\nolimits_{i=1}^{q}T^2\mathcal{L}_\text{KL}\left(\sigma\left(\frac{\mathbf{h}_i^\text{readout}}{T}\right), \sigma\left(\frac{\mathbf{z}_i^\text{readout}}{T}\right)\right),
\end{equation}
where $\mathcal{L}_\text{KL}$ denotes the Kullback-Leibler (KL) divergence loss~\cite{NIPS2003_0abdc563}, $\sigma$ denotes the softmax layer, and $T$ is a hyperparameter controlling the softening effect. We set $T$ to 4 throughout this paper, which is a common setting in existing methods~\cite{Chen_Mei_Zhang_Wang_Wang_Feng_Chen_2021, shu2021channel}. In Eq. (\ref{eq7}), $\mathbf{h}_i^\text{readout}$ and $\mathbf{z}_i^\text{readout}$ denote the atom-level and motif-level molecular representations of the $i$-th molecule, respectively. By optimizing $\mathcal{L}_\text{align}$, AMCT can further enhance the molecular representation capability.
\subsubsection{Motif Contrastive Loss}
Since atom-motif alignment loss is performed intra-molecule and it constrains the consistency between atoms and motifs in the same molecule, we naturally seek to perform inter-molecule contrast and investigate the consistency across different molecules. Given that identical motifs across different molecules exhibit similar chemical properties~\cite{hawker2005convergence, smith2018infrared}, it is expected that they should have similar representations across all molecules. To achieve this, we propose the motif contrastive loss, which maximizes the representation agreement of identical motifs across different molecules. Meanwhile, the representations of motifs belonging to different classes are pulled away. Specifically, the proposed motif contrastive loss can be defined as
\begin{equation}\label{eq8}
\mathcal{L}_{\text{contrastive}}^{\text{motif}}(\mathbf{Z}_{i:}^{\text{motif}})=\log \frac{\sum\nolimits_{k=1}^{l}{{\mathds{1}_{[{{y}_{i}}={{y}_{k}}]}}{{e}^{\left\langle {{\mathbf{Z}}_{i:}^{\text{motif}},{{\mathbf{Z}}_{k:}^{\text{motif}}}} \right\rangle }}}}{\sum\nolimits_{j=1}^{l}{{{e}^{\left\langle {{\mathbf{Z}}_{i:}^{\text{motif}}},{{\mathbf{Z}}_{j:}^{\text{motif}}} \right\rangle }}}},
\end{equation}
\begin{equation}\label{eq9}
\mathcal{L}_{\text{contrastive}}=-\frac{1}{l}\sum\nolimits_{i=1}^{l}{\mathcal{L}_{\text{contrastive}}^{\text{motif}}(\mathbf{Z}_{i:}^{\text{motif}})},
\end{equation}
where $l$ is the number of motifs in a batch of $q$ molecules, $\mathbf{Z}_{i:}^{\text{motif}}$, $\mathbf{Z}_{j:}^{\text{motif}}$, and $\mathbf{Z}_{k:}^{\text{motif}}$ denote the $i$-th, $j$-th, and $k$-th rows of $\mathbf{Z}^{\text{motif}}$, respectively. In Eq. (\ref{eq8}), the symbol $\langle \cdot \rangle$ expresses the inner product, and $\mathds{1}_{[\cdot]}$ is an indicator function which equals 1 if the argument inside the bracket holds, and 0 otherwise. In Eq. (\ref{eq8}), $y_{i}$ and $y_{k}$ are the motif labels of $\mathbf{Z}_{i:}^{\text{motif}}$ and $\mathbf{Z}_{k:}^{\text{motif}}$, respectively. In the proposed motif contrastive loss, the motifs belonging to identical/different classes are regarded as positive/negative pairs. By optimizing $\mathcal{L}_{\text{contrastive}}$, our proposed AMCT not only ensures the consistency in motif representations, but also improves the discriminating ability of learned motif representations.

\subsection{Property-Aware Decoding}
A good decoding process is also important to obtain the reliable representation. However, most existing GT models equally decode all properties without considering their importances. In contrast, we propose the property-aware decoding. Specifically, we first obtain property embeddings and then use the decoder to extract property-aware molecular representations. Finally, predicted results are obtained after the linear projection.

\subsubsection{Property Embedding}
Similar to motif embeddings, property embeddings are obtained via neural embedding vectors. Therefore, when we are given a set of molecular properties, the property embedding for the $i$-th molecular property category (\textit{e.g.}, toxicity) is denoted as $\mathbf{p}_i \in \mathbb{R}^{d}$, where $i=1,\ldots,c$ (here $c$ is the number of molecular property categories), and $\mathbf{p}_i$ is subsequently fed into the decoder.

\subsubsection{Decoder and Linear Layer}
The decoder (see Fig.~\ref{fig2}(g)) is designed to extract property-aware molecular representations. Given an input sequence of $\mathbf{p}_1, \ldots, \mathbf{p}_c$, we can represent them as a matrix $\mathbf{P} \in \mathbb{R}^{c \times d}$ with each row corresponding to a $\mathbf{p}_i$ ($i=1,\ldots, c$). The matrix $\mathbf{P}$ is first fed into the MHA, by which we obtain $\mathbf{P}^{(1)} \in \mathbb{R}^{c \times d}$. Next, $\mathbf{P} + \mathbf{P}^{(1)}$ is operated by a layer normalization operation, by which we obtain $\mathbf{P}^{(2)}$. To identify the motifs that are critical in deciding the properties of each molecule, we construct a property-aware attention mechanism. Specifically, we employ a cross-attention module, which uses property embeddings $\mathbf{P}^{(2)}$ as queries and motif representations $\mathbf{Z}^{\text{motif}}$ as keys and values. To summarize, this process is computed as
\begin{equation}\label{eq6}
    \left\{
    \begin{aligned}
    & \text{PAware}(\mathbf{Z}^{\text{motif}}, \mathbf{P}^{(2)})=\text{Con}(\text{head}_1^\text{cross}, \ldots, \text{head}_h^\text{cross})\mathbf{W}^{\mathbf{O}}, \\
    & \text{head}_i^\text{cross}=\text{croatt}(\mathbf{P}^{(2)}\mathbf{W}^{\mathbf{Q}}_i, \mathbf{Z}^{\text{motif}}\mathbf{W}^{\mathbf{K}}_i, \mathbf{Z}^{\text{motif}}\mathbf{W}^{\mathbf{V}}_i), \\
    & \text{croatt}(\mathbf{Q}, \mathbf{K}, \mathbf{V})=\text{crossweight}(\mathbf{Q}, \mathbf{K})\mathbf{V}, \\
    & \text{crossweight}(\mathbf{Q}, \mathbf{K})=\text{softmax}\left(\mathbf{Q}\mathbf{K}^{\top }/\sqrt{d_k}\right),
    \end{aligned}
    \right.
\end{equation}
where `$\text{PAware}$' denotes the property-aware attention mechanism, `$\text{croatt}$' denotes the computation of the cross-attention module, and `$\text{crossweight}$' denotes the computation of cross-attention weights. The matrix of cross-attention weights is denoted as $\mathbf{A} \in \mathbb{R}^{c \times m}$, which indicates the strength of interactions between property embeddings and motif representations. Therefore, a motif with larger cross-attention weight is considered to have a greater contribution to the molecular property. After training, we normalize the elements of $\mathbf{A}$ to the range $[0, 1]$. We then identify the motifs that are critical in deciding the properties of each molecule by considering the cross-attention weights that exceed the threshold $\alpha$, where $\alpha \in [0,1]$ is a hyperparameter. By sequentially passing through $L$ layers, we acquire the property-aware molecular representations. After the linear projection, the predicted results $\mathbf{o} \in \mathbb{R}^{c}$ are finally obtained.

\subsection{Model Training}
Since the supervised loss depends on the specific prediction objective, we adopt the cross-entropy loss for classification tasks, while we use the squared error loss for regression tasks. Given a batch of $q$ molecules, we employ two supervised loss functions, where one for the matrix of predicted results $\mathbf{O} \in \mathbb{R}^{q \times c}$, and the other for atom-level molecular representations $\mathbf{H}^{\text{readout}}$. Since $\mathbf{H}^{\text{readout}} \in \mathbb{R}^{q \times d}$, we use a linear layer to obtain $\mathbf{H}^{\text{linear}} \in \mathbb{R}^{q \times c}$. The overall loss of our proposed AMCT is shown in Eq. (\ref{eq10}), which assembles the supervised loss $\mathcal{L}_{\text{sup}}(\mathbf{O})$, the supervised loss $\mathcal{L}_{\text{sup}}(\mathbf{H}^{\text{linear}})$, the atom-motif alignment loss $\mathcal{L}_\text{align}$, and the motif contrastive loss $\mathcal{L}_{\text{contrastive}}$, namely,
\begin{equation}\label{eq10}
\mathcal{L}=\mathcal{L}_{\text{sup}}(\mathbf{O}) + \mathcal{L}_{\text{sup}}(\mathbf{H}^{\text{linear}}) + \lambda_\text{a}\mathcal{L}_\text{align} + \lambda_{\text{b}}\mathcal{L}_{\text{contrastive}},
\end{equation}
where $\lambda_\text{a} >0$ and $\lambda_{\text{b}} > 0$ are hyperparameters adjusting the impacts of $\mathcal{L}_\text{align}$ and $\mathcal{L}_{\text{contrastive}}$, respectively. In Eq.~(\ref{eq10}), if we are given a classification task, then the two loss functions are
\begin{equation}\label{eq11_1}
\mathcal{L}_{\text{sup}}(\mathbf{O})=-\frac{1}{qc}\sum\nolimits_{i=1}^{q}\sum\nolimits_{j=1}^{c} \mathbf{Y}_{ij} \ln \mathbf{O}_{ij},
\end{equation}
and
\begin{equation}\label{eq11_2}
\mathcal{L}_{\text{sup}}(\mathbf{H}^{\text{linear}})=-\frac{1}{qc}\sum\nolimits_{i=1}^{q}\sum\nolimits_{j=1}^{c} \mathbf{Y}_{ij} \ln \mathbf{H}_{ij}^{\text{linear}},
\end{equation}
respectively. If we are given a regression task, then they are
\begin{equation}\label{eq12_1}
\mathcal{L}_{\text{sup}}(\mathbf{O})=\frac{1}{qc}\sum\nolimits_{i=1}^{q}\sum\nolimits_{j=1}^{c} (\mathbf{Y}_{ij} - \mathbf{O}_{ij})^2,
\end{equation}
and
\begin{equation}\label{eq12_2}
\mathcal{L}_{\text{sup}}(\mathbf{H}^{\text{linear}})=\frac{1}{qc}\sum\nolimits_{i=1}^{q}\sum\nolimits_{j=1}^{c} (\mathbf{Y}_{ij} - \mathbf{H}_{ij}^{\text{linear}})^2,
\end{equation}
respectively, where $\mathbf{Y} \in \mathbb{R}^{q \times c}$ is the ground truth matrix.

\section{Experiments}
To validate the effectiveness of our proposed AMCT, we perform extensive experiments on seven widely used benchmark datasets. These datasets include both classification and regression tasks, which are all obtained from the OGB~\cite{hu2020open}. The statistical information and descriptions of these benchmark datasets are included in the supplementary material. Here we use the Area Under the ROC curve (AUC) as the evaluation metric (higher values are better) for the classification tasks, and use the Root Mean Square Error (RMSE) as the evaluation metric (lower values are better) for the regression task. To ensure a fair comparison, we calculate the mean AUC, mean RMSE, and the corresponding standard deviation over ten independent runs for each method on each dataset. The paired $t$-test with a significance level 0.05 is adopted for statistical significance test. We compare AMCT with several baseline methods, and their details can be found in the supplementary material.

\subsection{Visualization Results}
\begin{figure}[t]
  \centering
  \includegraphics[width=1\columnwidth]{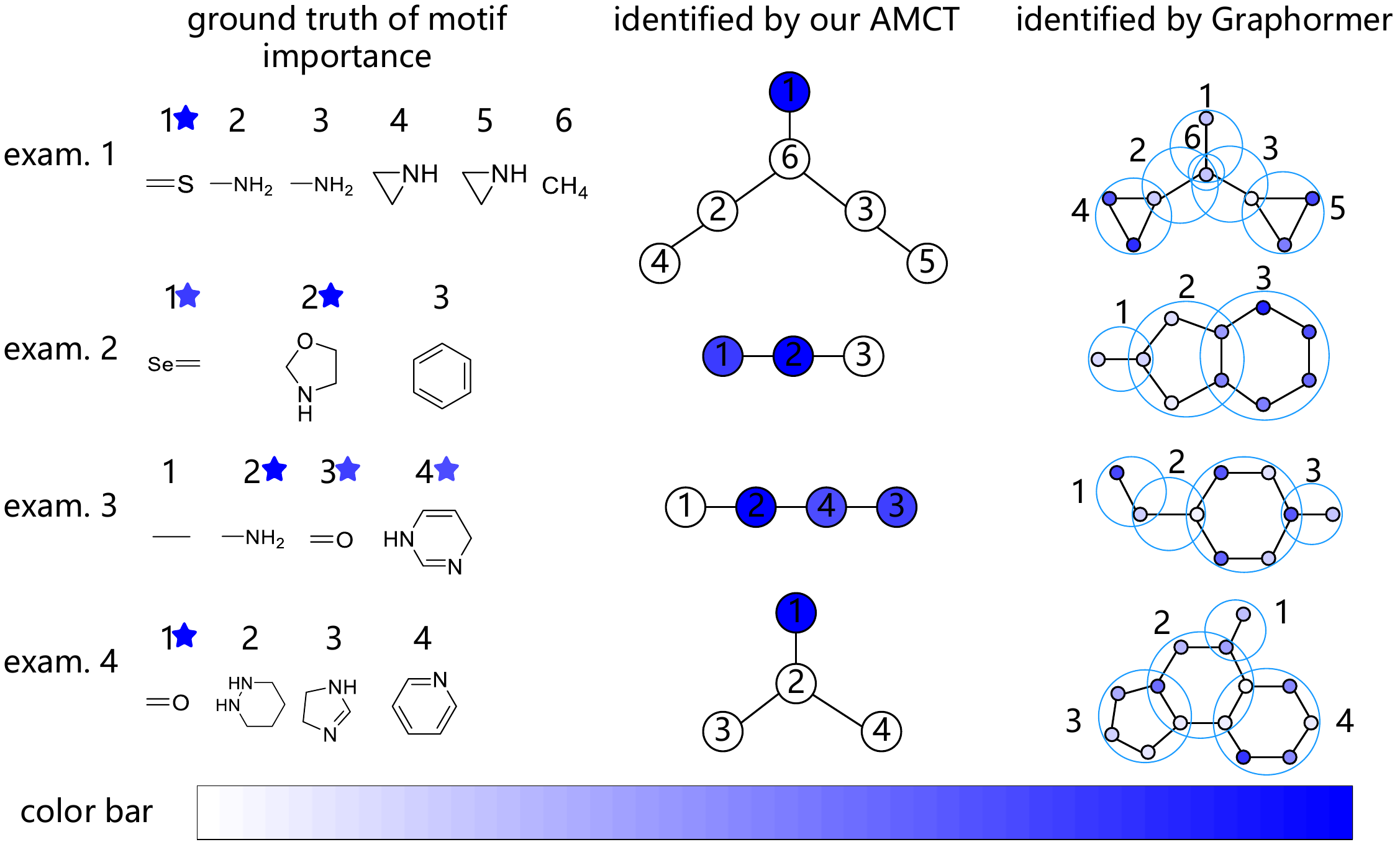}
  \caption{Visualization of identified motifs of four molecules in the \textit{HIV} dataset. Their IDs in the dataset are 265, 2444, 2547, and 6139, respectively. The blue stars in the first column indicate the motifs that are considered important in chemistry for influencing molecular properties. The blue hollow circles in the last column contain the atoms in the corresponding motifs, and the numbers around them are their corresponding motif IDs.}
  \label{fig5}
\end{figure}

To analyze whether the identified motifs are critical in deciding the properties of each molecule by using the cross-attention weights, we visualize the motifs identified by our proposed AMCT (the threshold $\alpha$ is set to 0.5) in the \textit{HIV} dataset, which is shown in Fig.~\ref{fig5}. Meanwhile, we also visualize the atom attention weights learned from Graphormer. It is noteworthy that these four molecules are all selected from the test set of the \textit{HIV} dataset. The task of the \textit{HIV} dataset is to predict whether a molecule will inhibit the HIV replication or not. The bluer the circles in Fig.~\ref{fig5} are, the larger the attention weights are.

In the second example, we can observe that `\ce{=S}' has the highest cross-attention weight in the motifs identified by AMCT, while other motifs such as `\ce{CH4}' have the lowest cross-attention weights. This makes sense because `\ce{=S}' is electrophilic, allowing it to interact with nucleophilic sites in RNA polymerase or RNA molecules. This interaction could alter the structure of RNA~\cite{ravichandran2008comparative}, which is the genetic material of HIV. Therefore, `\ce{=S}' has the potential to inhibit the HIV replication. However, the attention weight of the sulfur atom obtained from Graphormer is small, which means that Graphormer does not really identify motifs that are critical in deciding molecular properties. This is because Graphormer only considers the atom-level interactions but neglects the motif-level interactions. In contrast, owning to our proposed property-aware attention mechanism, the identified motifs are critical to the properties of the molecule. Due to space limitations, additional visualizations are provided in the supplementary material.

\subsection{Results of the Classification and Regression Tasks}
Tab.~\ref{table1} and Tab.~\ref{table2} show the AUCs and RMSEs of different methods on seven datasets for classification and regression tasks, where the best record on each dataset has been highlighted in bold.  As GCN~\cite{kipf2017semisupervised}, GIN~\cite{xu2018how}, and GSN~\cite{9721082} all use the `virtual node' technique~\cite{pmlrv70gilmer17a}, they are represented by `GCN+VN', `GIN+VN', and `GSN+VN' in Tab.~\ref{table2}, respectively. Notably, in Tab.~\ref{table1}, Graphormer and Molformer are both GT-based methods that achieve strong performances on six datasets, which is due to their high reliability in characterizing the latent relationship among atoms of each molecule. Our proposed AMCT not only explores the atom-level interactions but also considers the motif-level interactions, which are ignored by other GT-based methods. Meanwhile, AMCT enriches the self-supervisory signals by contrasting atom-level and motif-level representations, which is helpful to recognize critical patterns hidden in motifs. Consequently, our proposed AMCT consistently surpasses other GT-based methods and achieves the best performance among all compared methods on all datasets.

\begin{table*}[htb]
    \scriptsize
    \centering
    \caption{Experimental results of compared methods on six datasets for classification and regression tasks. The best records in each column are \textbf{bolded}. The `\checkmark' denotes that our AMCT is significantly better than the competitor by paired $t$-test.}
    \label{table1}
    \renewcommand{\arraystretch}{0.8} 
    \scalebox{1}{
    \begin{tabular}{l|lllll|l}
    \hline
               & \multicolumn{5}{c|}{Classification Tasks (AUC $\uparrow$)}                           & Regression Task (RMSE $\downarrow$) \\ \hline
    \multicolumn{1}{c|}{Methods}    & \multicolumn{1}{c}{\textit{ToxCast}}     & \multicolumn{1}{c}{\textit{Tox21}}       & \multicolumn{1}{c}{\textit{BBBP}}        & \multicolumn{1}{c}{\textit{BACE}}        & \multicolumn{1}{c|}{\textit{SIDER}}       & \multicolumn{1}{c}{\textit{ESOL}}            \\ \hline
    TF-Robust~\cite{ramsundar2015massively}  & 0.585$\pm$0.031 \checkmark & 0.698$\pm$0.012 \checkmark & 0.860$\pm$0.087 \checkmark & 0.824$\pm$0.022 \checkmark & 0.607$\pm$0.033            & \qquad1.722$\pm$0.038 \checkmark    \\ \hline
    Weave~\cite{kearnes2016molecular}      & 0.678$\pm$0.024 \checkmark & 0.741$\pm$0.044 \checkmark & 0.837$\pm$0.065 \checkmark & 0.791$\pm$0.008 \checkmark & 0.543$\pm$0.034 \checkmark & \qquad1.158$\pm$0.055 \checkmark    \\
    GCN~\cite{kipf2017semisupervised}        & 0.650$\pm$0.025 \checkmark & 0.772$\pm$0.041 \checkmark & 0.877$\pm$0.036 \checkmark & 0.854$\pm$0.011 \checkmark & 0.593$\pm$0.035 \checkmark & \qquad1.068$\pm$0.050 \checkmark    \\
    SchNet~\cite{NIPS2017_303ed4c6}     & 0.679$\pm$0.021 \checkmark & 0.767$\pm$0.025 \checkmark & 0.847$\pm$0.024 \checkmark & 0.750$\pm$0.033 \checkmark & 0.545$\pm$0.038 \checkmark & \qquad1.045$\pm$0.064 \checkmark    \\
    MGCN~\cite{lu2019molecular}       & 0.663$\pm$0.009 \checkmark & 0.707$\pm$0.016 \checkmark & 0.850$\pm$0.064 \checkmark & 0.734$\pm$0.030 \checkmark & 0.552$\pm$0.018 \checkmark & \qquad1.266$\pm$0.147 \checkmark     \\ 
    MGSSL~\cite{zhang2021motif}       & 0.638$\pm$0.003 \checkmark & 0.764$\pm$0.004 \checkmark & 0.705$\pm$0.011 \checkmark & 0.797$\pm$0.008 \checkmark & 0.605$\pm$0.007 \checkmark & \qquad1.179$\pm$0.008 \checkmark     \\ 
    GREA-GCN~\cite{liu2022graph}       & 0.658$\pm$0.006 \checkmark & 0.785$\pm$0.008 \checkmark & 0.679$\pm$0.018 \checkmark & 0.732$\pm$0.035 \checkmark & 0.569$\pm$0.023 \checkmark & \qquad0.897$\pm$0.036 \checkmark     \\ \hline
    MAT~\cite{maziarka2020molecule}        & 0.678$\pm$0.009 \checkmark & 0.785$\pm$0.011 \checkmark & 0.737$\pm$0.009 \checkmark & 0.846$\pm$0.025 \checkmark & 0.597$\pm$0.012 \checkmark & \qquad0.838$\pm$0.014 \checkmark     \\
    R-MAT~\cite{maziarka2021relative}      & 0.685$\pm$0.009 \checkmark & 0.791$\pm$0.009 \checkmark & 0.745$\pm$0.010 \checkmark & 0.858$\pm$0.041 \checkmark & 0.600$\pm$0.013 \checkmark & \qquad0.833$\pm$0.015 \checkmark     \\
    Graphormer~\cite{NEURIPS2021_f1c15925} & 0.703$\pm$0.010 \checkmark & 0.790$\pm$0.011 \checkmark & 0.917$\pm$0.010 \checkmark & 0.860$\pm$0.013 \checkmark & 0.616$\pm$0.010 \checkmark & \qquad0.829$\pm$0.014 \checkmark     \\
    Molformer~\cite{Wu_Fang_2023}  & 0.691$\pm$0.012 \checkmark & 0.783$\pm$0.012 \checkmark & 0.918$\pm$0.008 \checkmark & 0.881$\pm$0.014 \checkmark & 0.605$\pm$0.011 \checkmark & \qquad0.848$\pm$0.013 \checkmark     \\ \hline
    AMCT     & \textbf{0.715$\pm$0.011} & \textbf{0.804$\pm$0.010} & \textbf{0.927$\pm$0.008} & \textbf{0.895$\pm$0.012} & \textbf{0.633$\pm$0.016} & \qquad\textbf{0.815$\pm$0.010}     \\ \hline
    \end{tabular}
    }
\end{table*}

\begin{table*}[htb]
    \scriptsize
    \centering
    \caption{Ablation study of two losses on seven datasets. The best record in each column are \textbf{bolded}. The `\checkmark' denotes that our AMCT is significantly better than the reduced models by paired $t$-test.}
    \label{table3}
    \renewcommand{\arraystretch}{0.8} 
    \scalebox{1}{
    \begin{tabular}{l|llllll|l}
    \hline
    & \multicolumn{6}{c|}{Classification Tasks (AUC $\uparrow$)}                                   & Regression Task (RMSE $\downarrow$) \\ \hline
    \multicolumn{1}{c|}{Methods}            & \multicolumn{1}{c}{\textit{ToxCast}}     & \multicolumn{1}{c}{\textit{Tox21}}       & \multicolumn{1}{c}{\textit{BBBP}}        & \multicolumn{1}{c}{\textit{BACE}}        & \multicolumn{1}{c}{\textit{SIDER}}       & \multicolumn{1}{c|}{\textit{HIV}}         & \multicolumn{1}{c}{\textit{ESOL}}                   \\ \hline
    AMCT (w/o ALoss) & 0.699$\pm$0.009 \checkmark & 0.787$\pm$0.014 \checkmark & 0.909$\pm$0.014 \checkmark & 0.881$\pm$0.013 \checkmark & 0.614$\pm$0.018 \checkmark & 0.801$\pm$0.014 \checkmark & \qquad0.842$\pm$0.021 \checkmark            \\
    AMCT (w/o CLoss) & 0.695$\pm$0.010 \checkmark & 0.785$\pm$0.017 \checkmark & 0.907$\pm$0.016 \checkmark & 0.874$\pm$0.017 \checkmark & 0.612$\pm$0.017 \checkmark & 0.797$\pm$0.017 \checkmark & \qquad0.848$\pm$0.034 \checkmark            \\

    AMCT (w/o PAware) & 0.686$\pm$0.009 \checkmark & 0.778$\pm$0.011 \checkmark & 0.863$\pm$0.015 \checkmark & 0.862$\pm$0.014 \checkmark & 0.591$\pm$0.015 \checkmark & 0.776$\pm$0.013 \checkmark & \qquad0.896$\pm$0.024 \checkmark            \\ \hline

    AMCT             & \textbf{0.715$\pm$0.011} & \textbf{0.804$\pm$0.010} & \textbf{0.927$\pm$0.008} & \textbf{0.895$\pm$0.012} & \textbf{0.633$\pm$0.016} & \textbf{0.813$\pm$0.006} & \qquad\textbf{0.815$\pm$0.010}            \\ \hline
    \end{tabular}
    }
\end{table*}

\begin{table}[]
    \scriptsize
    \centering
    \caption{AUC of different methods on the \textit{HIV} dataset for the classification task. The best record is \textbf{bolded}. The `\checkmark' denotes that our AMCT is significantly better than the competitor by paired $t$-test.}
    \label{table2}
    \renewcommand{\arraystretch}{0.8} 
    \scalebox{1}{
    \begin{tabular}{l|l}
    \hline
    \multicolumn{1}{c|}{Methods}    & \multicolumn{1}{c}{Classification Task (AUC $\uparrow$)}         \\ \hline
    GCN+VN~\cite{kipf2017semisupervised}     & \qquad0.760$\pm$0.012 \checkmark \\
    GIN+VN~\cite{xu2018how}     & \qquad0.771$\pm$0.015 \checkmark \\
    PNA~\cite{NEURIPS2020_99cad265}        & \qquad0.791$\pm$0.013 \checkmark \\
    MGSSL~\cite{zhang2021motif}        & \qquad0.795$\pm$0.011 \checkmark \\
    DGN~\cite{pmlrv139beaini21a}        & \qquad0.797$\pm$0.010 \checkmark \\
    GREA-GCN~\cite{liu2022graph}        & \qquad0.762$\pm$0.019 \checkmark \\
    HM-GNN~\cite{pmlrMolecular}     & \qquad0.790$\pm$0.009 \checkmark \\ 
    MolR-GCN~\cite{wang2022chemicalreactionaware}     & \qquad0.802$\pm$0.024 \\ 
    GSN+VN~\cite{9721082}     & \qquad0.780$\pm$0.001 \checkmark \\ \hline
    Graphormer~\cite{NEURIPS2021_f1c15925} & \qquad0.805$\pm$0.005 \checkmark \\ 
    SAN~\cite{NEURIPS2021_b4fd1d2c}        & \qquad0.779$\pm$0.002 \checkmark \\
    GPS~\cite{NEURIPS2022_5d4834a1}        & \qquad0.788$\pm$0.010 \checkmark \\ \hline
    AMCT     & \qquad\textbf{0.813$\pm$0.006} \\ \hline
    \end{tabular}
    }
\end{table}

\subsection{Ablation Study}
\begin{figure}[]
  \centering
  \includegraphics[width=1\columnwidth]{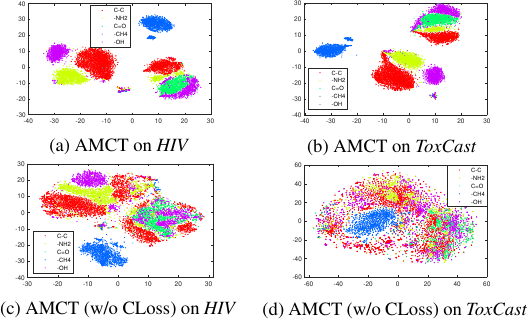}
  \caption{The t-SNE visualizations of motif representations obtained from different models on two datasets.}
  \label{figtsne}
\end{figure}

Our proposed AMCT employs atom-motif alignment loss and motif contrastive loss to enrich the self-supervisory signals. In addition, we use the property-aware attention mechanism to improve the reliability of MPP. To shed light on the contributions of these components, we report the experimental results of AMCT when each of these components is removed on the seven datasets, which are shown in Tab.~\ref{table3}. For simplicity, `AMCT (w/o ALoss)', `AMCT (w/o CLoss)', and `AMCT (w/o PAware)' denote the reduced models by removing $\mathcal{L}_\text{align}$, $\mathcal{L}_{\text{contrastive}}$, and property-aware attention mechanism, respectively. It can be observed that the performance decreases when either $\mathcal{L}_\text{align}$ or $\mathcal{L}_{\text{contrastive}}$ is removed, showing that both loss functions contribute significantly to satisfactory performance. In particular, AMCT is able to significantly improve the performance by using the motif contrastive loss. For example, AUC can be decreased by more than 2 percentage on the \textit{BACE} dataset without $\mathcal{L}_{\text{contrastive}}$. Moreover, the performance decreases when removing the property-aware attention mechanism, which validates that it can help obtain reliable molecular representations. Furthermore, we select five types of motifs and then visualize their motif representations obtained from AMCT and `AMCT (w/o CLoss)' on the \textit{HIV} and \textit{ToxCast} datasets via using the t-SNE method~\cite{van2008visualizing}, respectively. As shown in Fig.~\ref{figtsne}, the 2D projections of the motif representations obtained from AMCT (see Fig.~\ref{figtsne}(a) and Fig.~\ref{figtsne}(b)) show more compact clusters when compared to `AMCT (w/o CLoss)' (see Fig.~\ref{figtsne}(c) and Fig.~\ref{figtsne}(d)). Therefore, it demonstrates that our proposed motif contrastive loss is beneficial for promising performances.

\subsection{Sensitivity Analysis}
\begin{figure}
    \centering
    \begin{subfigure}{0.23\textwidth}
      \includegraphics[width=\linewidth]{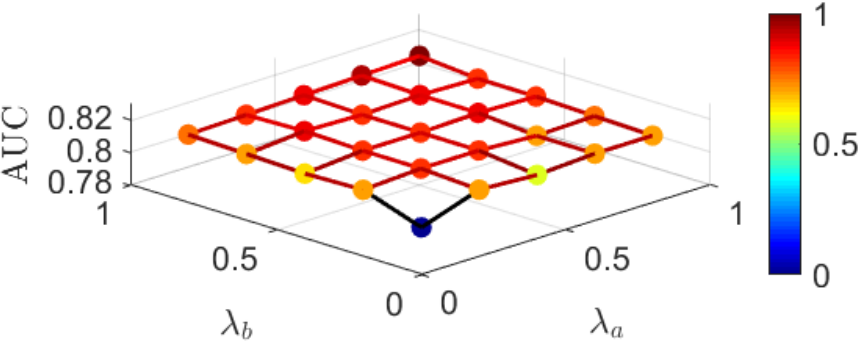}
      \caption{\textit{HIV} dataset}
      \label{fig6_1}
    \end{subfigure}
    \begin{subfigure}{0.23\textwidth}
      \includegraphics[width=\linewidth]{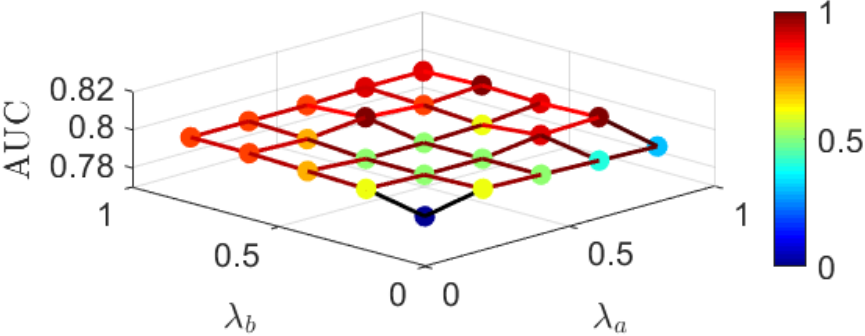}
      \caption{\textit{Tox21} dataset}
      \label{fig6_2}
    \end{subfigure}
    \begin{subfigure}{0.23\textwidth}
      \includegraphics[width=\linewidth]{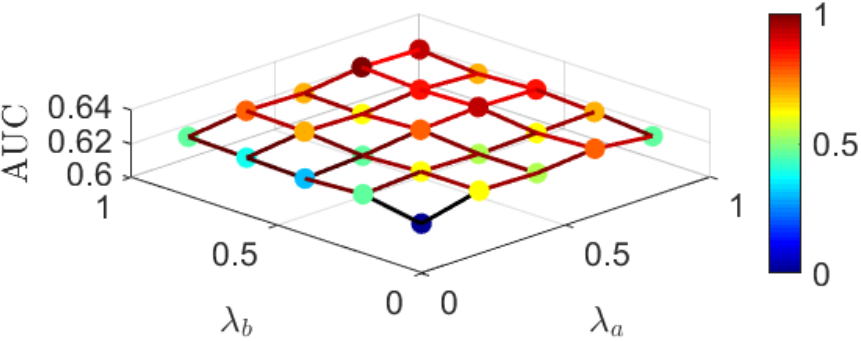}
      \caption{\textit{SIDER} dataset}
      \label{fig6_3}
    \end{subfigure}
    \begin{subfigure}{0.23\textwidth}
      \includegraphics[width=\linewidth]{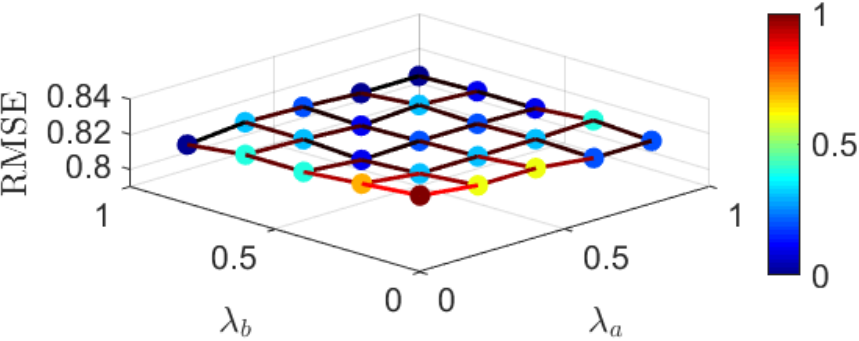}
      \caption{\textit{ESOL} dataset}
      \label{fig6_4}
    \end{subfigure}
    \caption{The sensitivity analyses of the hyperparameters $\lambda_\text{a}$ and $\lambda_{\text{b}}$ on four datasets.}
    \label{fig6}
  \end{figure}

Since there are two hyperparameters $\lambda_\text{a}$ and $\lambda_{\text{b}}$ in Eq.~(\ref{eq10}), we analyze the sensitivity of the performance with different hyperparameter settings. Specifically, we conduct the sensitivity analyses on four datasets (\textit{i.e.}, the \textit{HIV}, \textit{Tox21}, \textit{SIDER}, and \textit{ESOL} datasets), which are shown in Fig.~\ref{fig6}. We can observe that the performance of AMCT is relatively stable in terms of AUC and RMSE metrics. Therefore, the performance of our proposed AMCT is actually insensitive to the choice of hyperparameters, and thus the parameters of our method can be easily tuned in practical uses. 

\section{Conclusion}
In this paper, we proposed a novel Atom-Motif Contrastive Transformer (AMCT) for MPP, which simultaneously considers the atom-level and motif-level interactions within the molecule. The main advantage of our method is that it exploits the critical contrast information within atoms and motifs, and thus successfully building supervisory signals for reliable model training. The proposed contrastive transformer was further integrated with a new property-aware attention mechanism, so that we could finally identify the critical motifs in deciding the molecular properties. Due to the effectiveness of the above contrastive transformer learning as well as the compatibility of the new attention block, our method achieved significantly better results than eleven representative approaches on seven benchmark datasets.

\bibliography{cite}

\end{document}